\documentclass[cameraready]{Interspeech}

\title{Light-weight Pronunciation Assessment via\\Discrete Speech Token Surprisal}

\author[affiliation=, orcid=0009-0005-6804-6181]{Syeda Faiza Ahmed}{Sara}
\author[affiliation=, orcid=0000-0002-1331-2543, correspondingauthor]{Shammur Absar}{Chowdhury}

\address{
     Qatar Computing Research Institute, Doha, Qatar
}

\email{syeda.faiza.ahmed@gmail.com, shchowdhury@hbku.edu.qa}

\keywords{Pronunciation Assessment, Discrete speech tokens, Token surprisal,
          Self-supervised learning, Unsupervised, Computer-assisted language learning}

\usepackage{comment}
\usepackage{booktabs}
\usepackage{amsmath}
\usepackage{tabularx}
\usepackage{array}

\begin{document}

\maketitle

\begin{abstract}
Training automated pronunciation assessment often relies on labeled learner errors or non-native corpora that are costly to collect. We propose a lightweight framework trained only on native speech resources, operating unsupervised or lightly calibrated with a small set of scored utterances. At inference, learner speech is discretized with an SSL encoder and a K-means codebook. A token language model trained on native sequences computes surprisal where higher surprisal indicates phonotactic deviation. We add a transcript-guided Text2DUnit--DTW module that predicts native token sequences from reference text and aligns them to acoustic tokens to derive error-sensitive features. Surprisal and alignment features are fused via simple regression. On SpeechOcean762, PCC improves from 0.60 to 0.66 with transcript guidance, near supervised baselines. Cross-dataset evaluation on L2-ARCTIC shows consistent gains.
\end{abstract}


\section{Introduction}

Pronunciation assessment is central to computer-assisted language learning, yet
building reliable automatic systems remains difficult when labeled data are
scarce. The most widely used approach computes Goodness of Pronunciation (GoP)
scores from ASR acoustic models \cite{witt2000phone}, which requires forced
alignment and reference transcriptions. Regression-based approaches that map
learned representations to expert scores \cite{gong2022transformer} avoid
explicit phone-level modeling, but still require labeled non-native speech that
is expensive to collect and often domain-specific. These requirements limit
progress in settings where learner corpora are unavailable, including endangered
or low-resource varieties, specialized speaking styles such as liturgical
recitation, and many classroom scenarios.

At the same time, discrete speech tokens obtained by clustering self-supervised
representations have become a practical abstraction for speech processing
\cite{hsu2021hubert,babu2021xls}. Language models over token sequences capture
regularities related to phonotactics and have enabled applications such as speech recognition \cite{sukhadia2024}, speech generation \cite{lakhotia2021generative} and acoustic anomaly detection
\cite{han2025exploring}. This motivates a simple idea for pronunciation
assessment. If a model trained only on native speech learns what typical token
sequences look like, it should assign higher surprisal to token patterns that
deviate from native phonotactics, as often happens in non-native speech.

We build on this intuition and introduce a lightweight framework that supports unsupervised scoring with or without reference text, and can be lightly supervised when a small set of scored learner utterances is available, making it suitable for zero-resource settings.
First, we discretize speech using a frozen self-supervised encoder and a K-means codebook trained on native speech. We then score the resulting token sequences with a small n-gram language model, requiring no phoneme inventory, forced alignment, or mispronunciation labels.
When reference transcriptions are available, as in reading-based assessment, we
add a transcript-guided component. A compact seq2seq model predicts the
canonical native token sequence for the reference text, and dynamic time warping (DTW)
aligns it to the learner acoustic tokens. This yields fine-grained alignment features that directly compare expected and observed token patterns in discrete space, still without an ASR system or forced aligner.
\noindent Our main contributions are as follows.
\begin{itemize}[noitemsep,topsep=0pt,leftmargin=*,labelsep=.5em]
    \item We introduce a lightweight pronunciation assessment framework that supports
    unsupervised scoring with or without reference text, and optional light calibration
    with a small set of scored learner utterances.
    \item We propose native-trained discrete-token surprisal features that avoid phoneme
    inventories, forced alignment, and learner or mispronunciation labels.
    \item We add a transcript-guided Text2DUnit module and DTW alignment in discrete space to
    derive fine-grained mispronunciation features.
    \item We show that combining surprisal and alignment improves over audio-only scoring,
    exceeds prior zero-shot results on SpeechOcean762, and transfers to L2-ARCTIC.
\end{itemize}

\begin{figure}[t]
    \centering
    \includegraphics[width=\linewidth]{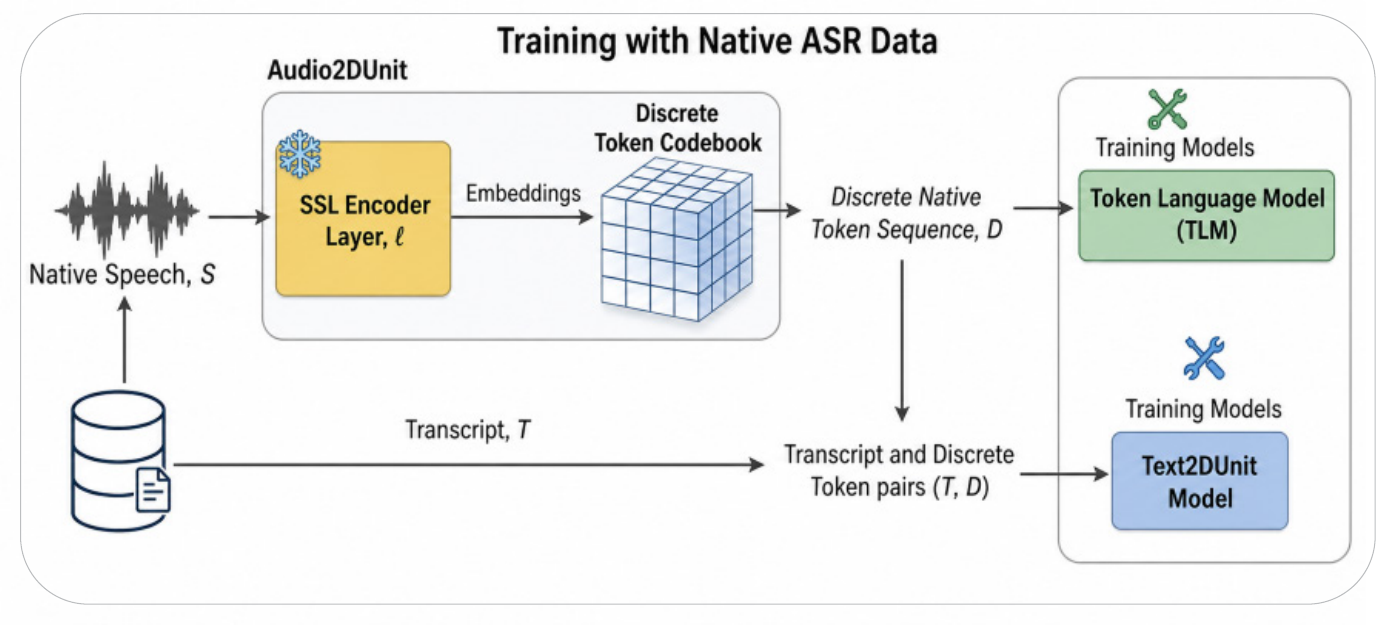}
    \vspace{-0.5cm}
    \caption{\textbf{Training overview.}
    Training uses only standard native speech (ASR) resources and requires no learner data,
    manual annotation, or forced alignment.}
    \label{fig:training}
    \vspace{-0.5cm}
\end{figure}

\begin{figure*}[t]
    \centering
    \includegraphics[width=0.8\textwidth]{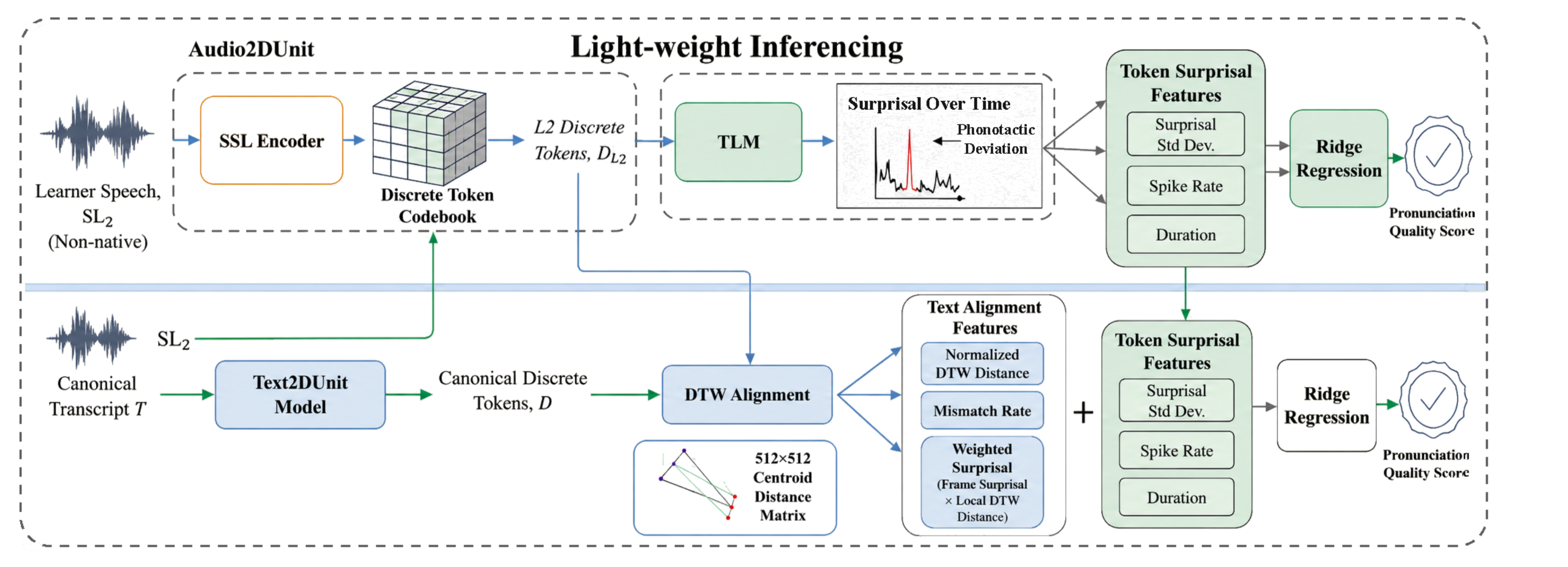}
    \vspace{-0.3cm}
    \caption{\textbf{Inference overview.}
    At inference, we compute audio-only surprisal features and optional transcript-guided DTW alignment features.
    A simple regressor can be trained with a small set of annotated learner samples, but the features also act as
    direct pronunciation quality indicators.}
    \label{fig:inference}
    \vspace{-0.5cm}
\end{figure*}

\section{Method}
\label{sec:method}



Figures~\ref{fig:training} and~\ref{fig:inference} illustrate the proposed pronunciation assessment framework. During training (Figure~\ref{fig:training}), we use only native speech to learn a discrete unit vocabulary and a native phonotactic prior through two modules: Audio2DUnit and the Token-level Language Model (TLM). We also train Text2DUnit to map reference transcripts into the same discrete unit space.

At inference (Figure~\ref{fig:inference}), learner speech is converted into discrete units using Audio2DUnit. We then extract two feature groups for regression: (i) \textbf{surprisal statistics}, which measure deviation from the native phonotactic prior, and (ii) \textbf{transcript-guided alignment}, which compares learner units with canonical text-derived units using Dynamic Time Warping (DTW). The following sections describe each module in the pipeline.

\vspace{-0.2cm}
\subsection{Audio-to-Discrete-Unit (Audio2DUnit)}
\label{sec:tokenization}

The Audio2DUnit module acts as the primary acoustic tokenizer, converting continuous 16\,kHz speech into discrete symbolic units. We first extract frame-level representations using a pretrained SSL encoder and then quantize these embeddings with a codebook of size $K$. The codebook is learned by applying K-means clustering to SSL features from a native speech corpus.
The K-means discretization provides a compact symbolic abstraction of the acoustic space, the resulting units are derived from rich SSL representations rather than phoneme labels. This allows them to capture salient temporal and acoustic patterns, including transitions, fluency, and prosodic structure. 
After training on native speech, the Audio2DUnit module is frozen to preserve a native-only phonotactic representation. At inference, each frame is mapped to its nearest centroid, producing a discrete token sequence $t_1, t_2, \ldots, t_N$.


\vspace{-0.2cm}
\subsection{Native Token Language Model (TLM)}

To characterize the phonotactic distribution of native speech, we train a Token Language Model (TLM) on the discrete sequences extracted by the Audio2DUnit module. We employ a n-gram model that estimates the conditional probability of each token given its local context.

\vspace{-0.2cm}
\subsection{Text-to-Discrete-Unit (Text2DUnit)}

The Text2DUnit module maps reference transcripts into the same discrete unit space as the learner's acoustic tokens. It uses character-level text representations, which make the model robust to spelling variation, orthographic differences, and out-of-vocabulary words. Trained on native text-audio pairs, it predicts the collapsed (``deduplicated'') discrete unit sequence produced by Audio2DUnit for the corresponding speech, producing text-derived canonical units that represent the expected native pronunciation.
As both Text2DUnit and Audio2DUnit now share the same discrete vocabulary, the system can directly compare the intended pronunciation with the learner's acoustic realization during assessment.

\vspace{-0.2cm}
\subsection{Pronunciation Assessment Module}
The final pronunciation score is predicted using a light-weight regression model trained on high-level feature descriptors. These features are derived from the learner's acoustic tokens by evaluating them against a native phonotactic prior (using TLM) and, optionally, a canonical reference sequence (DTW-alignment).

\vspace{-0.2cm}
\paragraph*{Surprisal Feature Extraction}
To quantify phonotactic deviation directly from the speech signal, we use the TLM to compute token-level surprisal~\cite{hale2001probabilistic, levy2008expectation}, a standard information-theoretic measure of contextual unpredictability. Let $T=(t_1,\ldots,t_N)$ denote the discrete unit sequence produced by the Audio2DUnit module. For each token $t_i$, surprisal is computed from its trigram context as: $S(t_i) = -\log_2 P(t_i \mid t_{i-2},, t_{i-1}) \quad \text{bits}$.
Higher surprisal indicates that a token is less predictable given its preceding context, and therefore reflects a stronger local phonotactic deviation.


We summarize the utterance-level surprisal profile using three descriptive features:
\begin{itemize}[noitemsep,topsep=0pt,leftmargin=*,labelsep=.5em]
\item \textbf{Surprisal Std.\ Dev.}: the standard deviation of ${S(t_i)}$, which captures variability in the surprisal profile and highlights localized surprisal spikes;
\item \textbf{Spike Rate}: the proportion of tokens whose surprisal exceeds the 90th-percentile threshold estimated from native speech, set to 9.0 bits;
\item \textbf{Duration}: the total count of discrete units in the utterance, used as a proxy for speaking rate and fluency.
\end{itemize}

\noindent We use surprisal standard deviation rather than mean surprisal because pronunciation errors are typically sparse and localized. Mean surprisal averages these deviations over the entire utterance and can therefore obscure short but important anomalies. In contrast, standard deviation captures the spikiness of the surprisal profile, making it better suited for detecting isolated phonotactic deviations against otherwise native-like token transitions.


\vspace{-0.2cm}
\paragraph*{Transcript-guided Alignment Features (Optional)}
When the reference text is available, we incorporate a second layer of features to capture text-specific signals. We use the Text2DUnit module to generate a canonical token sequence, $T_t$, from the text. We then perform DTW to align the learner's acoustic sequence ($T_a$) in the shared discrete symbolic space. Both the text and acoustic token sequences undergo consecutive-duplicate collapsing, while $T_t$ is collapsed by design during model training, for $T_a$ we independently collapse the sequence by de-duplicating the consecutive token labels and get $\hat{T_a}$.

\noindent\textbf{DTW alignment:} We align $T_t$ against $\hat{T_a}$ using DTW~\cite{giorgino2009computing}. The local cost at each alignment step $(i, j)$ is the $L^2$ distance between the K-means centroids corresponding to tokens $T_t[i]$ and $\hat{T_a}[j]$ in SSL embedding space: $\delta(i,j) = \lVert \mathbf{c}_{T_t[i]} - \mathbf{c}_{\hat{T_a}[j]} \rVert_2$
where $\{\mathbf{c}_k\}$ are the fitted K-means centroids. To avoid recomputing distances at every step, we precompute a $512 \times 512$ matrix $D$ with $D[i][j] = \lVert \mathbf{c}_i - \mathbf{c}_j \rVert_2$ before alignment. This centroid-based cost is acoustically meaningful: tokens corresponding to similar phonetic realizations have nearby centroids and thus low alignment cost, even when assigned different cluster indices. A binary match/mismatch cost would treat all substitutions as equally severe; centroid distance distinguishes acoustically close substitutions (e.g., a neighboring vowel cluster) from distant ones (e.g., a vowel replaced by a fricative cluster).
The raw DTW path cost is normalized by the path length $L$ (number of alignment steps) rather than by either sequence length individually, ensuring comparability across utterances of different durations. If the optimal path visits steps $(i_1, j_1), \ldots, (i_L, j_L)$, the normalized cost is: 

\begin{equation}
    \label{eq:dtw-distance}
  \footnotesize
    \text{DTW Distance} = \frac{1}{L} \sum_{\ell=1}^{L} D[i_\ell,\, j_\ell]
\end{equation}


Following, four features are derived from the alignment path and the frame-rate surprisal sequence. For features requiring frame-level resolution (mismatch surprisal std and weighted surprisal std), each raw frame in $a^d$ inherits the local DTW distance of its corresponding collapsed position.

\begin{description} [noitemsep,topsep=0pt,leftmargin=*,labelsep=.5em]
  \item[DTW Distance:] the normalized path cost defined in Eq.~\eqref{eq:dtw-distance}. High values indicate the learner's acoustic realization is, on average, far from the expected native token sequence after accounting for timing differences.

  \item[Token Mismatch Rate:] the fraction of alignment steps where the actual and predicted tokens differ, $a^d_c[j] \neq t^d[i]$. This provides a discrete measure of segmental substitution frequency at collapsed-token resolution.

  \item[Mismatch Surprisal Std:] the standard deviation of frame-level surprisal $S(t_i)$ restricted to frames whose collapsed position is flagged as a mismatch in the DTW alignment. This isolates surprisal variability at frames the system explicitly identifies as phonotactically inconsistent with the reference.

  \item[Weighted Surprisal Std:] the standard deviation of $S(t_i) \times (1 + \alpha \cdot \delta_i)$ computed at frame resolution, where $\delta_i$ is the local centroid distance inherited from the alignment and $\alpha = 0.5$. Frames that are simultaneously high-surprisal under the LM and acoustically distant from the expected token contribute disproportionately, amplifying the signal at the most probable mispronunciation sites.
\end{description}

\vspace{-0.2cm}
\subsection{Feature Combination} 
\label{sec:combination}

We combine TLM-derived surprisal features with transcript-guided alignment features from Text2DUnit--DTW using Ridge regression to predict a single pronunciation quality score. We evaluate three configurations: audio-only features, transcript-guided alignment features only, and their combination.



\section{Experiments}
\label{sec:experiments}

\subsection{Data}
\textbf{Training.}
The Audio2DUnit codebook and Token Language Model (TLM) are trained on LibriSpeech 960\,h of native English read speech~\cite{panayotov2015librispeech}.
The Text2DUnit model is trained on (transcript, collapsed-token) pairs extracted from the LibriSpeech 960-hour setup and tokens from the same Audio2DUnit K-means codebook, with a 90/10 random train/validation split.

\noindent\textbf{Evaluation.}
Primary evaluation uses the SpeechOcean762~\cite{zhang2021speechocean762} test split (2{,}500 utterances from 250 non-native Mandarin-English speakers), annotated at phoneme, word, and sentence level by five experts. 

\noindent\textbf{Generalization.}
Cross-dataset evaluation uses L2-ARCTIC~\cite{zhao2018l2arctic}, comprising 24 non-native English speakers each reading 1{,}150 identical sentences. Utterance-level pronunciation quality labels are obtained from Microsoft Azure Pronunciation Assessment pseudo-labels. Following, an expert manually went over for 10 speakers (1{,}351 utterances). We report correlation against four PA dimensions: AccuracyScore, FluencyScore, and PronScore (0--100).

\subsection{Training Parameters}
\label{subsec:training_params}

\textbf{Audio2DUnit.}
We extract 16\,kHz frame representations from HuBERT base~\cite{hsu2021hubert} Layer~9 and fit a K-means codebook with $K=512$ on 10{,}000 utterances sampled from the LibriSpeech 960-hour setup.
The encoder, layer, and vocabulary size are selected via an ablation sweeping four SSL encoders, layers $\in \{6, 9, 12\}$, and $K \in \{100, 256, 512, 1024, 2048\}$. 

\noindent\textbf{TLM.}
We train a 3-gram token LM on LibriSpeech token sequences produced by Audio2DUnit.

\noindent\textbf{Text2DUnit.} We use a CANINE-S \cite{clark2022canine} character encoder with frozen backbone
and LoRA adapters (rank $=32$, $\alpha=64$) on attention projections, and a
4-layer Transformer decoder ($d_{\text{model}}=768$, 8 heads,
$d_{\text{ff}}=2048$, dropout $=0.1$). The output vocabulary has 514 symbols (512 tokens plus EOS and PAD). Training uses AdamW (lr $=5\times10^{-5}$, weight decay $=0.01$), batch size 64, cosine schedule with 1{,}000-step warmup, fp16, and early stopping (patience $=7$).

\noindent\textbf{DTW and scoring.}
We align predicted and acoustic collapsed token sequences using DTW with centroid $L_2$ distance, using a precomputed $512\times512$ distance matrix. Weighted surprisal uses $\alpha=0.5$. For supervised calibration, we fit Ridge regression ($\alpha=1.0$, StandardScaler) on SpeechOcean762 and apply the same model to L2-ARCTIC without retraining.

\noindent\textbf{Evaluation metrics.}
For evaluating correlation, we opt for Pearson Correlation Coefficient (PCC) as the main measure for utterance level accuracy, fluency and prosody dimension.

\section{Results and Discussion}
\label{sec:results}

\noindent\textbf{Evaluation Settings}: For the study, we evaluate \textit{three inference settings} on SpeechOcean762. First, an \textbf{unsupervised} setting in which the proposed
features are used directly as pronunciation quality indicators without any
supervised regression model (see Table \ref{tab:features}). Second, an
\textbf{audio-only} setting (\textit{light-supervised}), where pronunciation features derived from token
surprisal statistics are mapped to scores using a lightweight regression
model (see Table \ref{tab:main}). Third, a \textbf{transcript-guided} (\textit{light-supervised}) setting that incorporates features derived from the reference transcript along with audio-based features followed by a regression head (Table \ref{tab:main}).

\noindent\textbf{Unsupervised Settings} Results in Table~\ref{tab:features} use each proposed feature directly as a pronunciation quality indicator, without any supervised regression model. In this setting, transcript-guided alignment signals are the strongest individual predictors, with DTW distance reaching $-0.633/-0.709/-0.707$ (Acc./Flu./Pros.) and mismatch rate showing comparable trends. Among audio-only signals, duration is most predictive, while surprisal-based features provide a complementary but weaker correlation, with surprisal standard deviation achieving $-0.316/-0.307/-0.341$. 

\noindent\textbf{Light-supervised Settings} Adding a light Ridge regressor 
improves calibration and feature fusion, reaching
$0.597/0.694/0.688$ with audio-only features and $0.661/0.763/0.753$ when
combining audio and transcript-guided features.

\noindent\textbf{Comparison with Existing Models.}
Table~\ref{tab:main} compares our method with supervised and label-free baselines
on SpeechOcean762. Supervised models such as HMamba perform best but rely on
substantial labeled learner data. In the label-free setting, our audio-only
system matches aMRT (Acc.~PCC$=0.597$ vs.\ 0.60) with a single forward pass.
DTW distance alone is a strong zero-shot indicator (Acc.~PCC$=0.633$), and adding
transcript-guided signals improves to Acc./Flu./Pros.\ PCC$=0.661/0.763/0.753$.

\begin{table}[t]
  \caption{PCC on SpeechOcean762 test. $\dagger$: from~\cite{liu2023zeroshot}.}
  \vspace{-0.3cm}
  \label{tab:main}
  \centering
  \footnotesize
  \setlength{\tabcolsep}{3pt}
  \renewcommand{\arraystretch}{0.90}
  \begin{tabularx}{\columnwidth}{@{}>{\raggedright\arraybackslash}Xccc@{}}
    \toprule
    Method & Acc. & Flu. & Pro. \\
    \midrule
    \multicolumn{4}{@{}c@{}}{\textit{Supervised}} \\ \midrule
    GoP~\cite{witt2000phone}$^\dagger$         & 0.64 & -- & -- \\
    DeepFeature$^\dagger$                     & 0.72 & -- & -- \\
    GOPT~\cite{gong2022transformer}$^\dagger$ & 0.74 & -- & -- \\
    MultiPA~\cite{chen2024multipa}            & 0.705 & 0.772 & 0.764 \\
    HMamba~\cite{chao2025towards}      & \textbf{0.807} & \textbf{0.848} & \textbf{0.843} \\
    \midrule

    \multicolumn{4}{@{}c@{}}{\textit{Zero-shot / unsupervised}} \\ \midrule
    non-reg GoP$^\dagger$                    & 0.57 & -- & -- \\
    Liu et al.~\cite{liu2023zeroshot} (aMRT) & 0.60 & -- & -- \\
    \midrule

    \multicolumn{4}{@{}c@{}}{\textit{Ours, trained with 100 hours}} \\ \midrule
    Transcript-only (DTW dist.)       & 0.611 & 0.664 & 0.668 \\
    Audio-only                        & 0.601 & 0.694 & 0.680 \\
    Audio + transcript-guided         & \textbf{0.668} & \textbf{0.757} & \textbf{0.748} \\
    \midrule

    \multicolumn{4}{@{}c@{}}{\textit{Ours, trained with 960 hours}} \\ \midrule
    Transcript-only (DTW dist.)       & 0.633 & 0.709 & 0.707 \\
    Audio-only                        & 0.597 & 0.694 & 0.688 \\
    Audio + transcript-guided         & \textbf{0.661} & \textbf{0.763} & \textbf{0.753} \\
    \bottomrule
  \end{tabularx}
  \vspace{-0.2cm}
\end{table}


\begin{table}[t]
  \caption{Feature-level PCC on SpeechOcean762. Source: audio (A) or transcript (T).}
  \vspace{-0.3cm}
  \label{tab:features}
  \centering
  \footnotesize
  \setlength{\tabcolsep}{2.4pt}
  \renewcommand{\arraystretch}{0.92}
  \begin{tabularx}{\columnwidth}{@{}>{\raggedright\arraybackslash}Xcccc@{}}
    \toprule
    Feature & Src & Acc. & Flu. & Pros. \\
    \midrule
    \multicolumn{5}{@{}c@{}}{\textit{Unsupervised, no learner corpus used in training}} \\ \midrule
    DTW distance           & T & $-$0.633 & $-$0.709 & $-$0.707 \\
    Token mismatch rate    & T & $-$0.621 & $-$0.688 & $-$0.690 \\
    Duration               & A & $-$0.534 & $-$0.649 & $-$0.625 \\
    Weighted surprisal std & T & $-$0.426 & $-$0.445 & $-$0.465 \\
    Surprisal std          & A & $-$0.316 & $-$0.307 & $-$0.341 \\
    Spike rate             & A & $-$0.272 & $-$0.264 & $-$0.304 \\
    Mismatch surprisal std & T & $-$0.192 & $-$0.173 & $-$0.204 \\
    \midrule

    \multicolumn{5}{@{}c@{}}{\textit{Ridge trained on SO762 train data}} \\ \midrule
    Ridge audio-only       & --  & 0.597 & 0.694 & 0.688 \\
    Ridge combined         & --  & \textbf{0.661} & \textbf{0.763} & \textbf{0.753} \\
    \bottomrule
  \end{tabularx}
  \vspace{-0.3cm}
\end{table}

  


\begin{table}[t]
  \caption{L2-ARCTIC PCC results. Utterance-level scores are averaged across speakers.}
  \vspace{-0.3cm}
  \label{tab:l2arctic_summary}
  \centering
  \footnotesize
  \setlength{\tabcolsep}{2.4pt}
  \renewcommand{\arraystretch}{0.92}
  \begin{tabularx}{\columnwidth}{@{}>{\raggedright\arraybackslash}Xccc@{}}
    \toprule
    Method & Acc. & Flu. & Pron. \\
    \midrule
    \multicolumn{4}{@{}c@{}}{\textit{Zero-shot features}} \\ \midrule
    DTW distance     & $-$0.500 & $-$0.385 & $-$0.512 \\
    Mismatch rate    & $-$0.492 & $-$0.353 & $-$0.502 \\
    \midrule

    \multicolumn{4}{@{}c@{}}{\textit{Ridge regression}} \\ \midrule
    Ridge (SO-train) & 0.506 & 0.492 & 0.526 \\
    Ridge (L2-train) & \textbf{0.527} & \textbf{0.519} & \textbf{0.557} \\
    \bottomrule
  \end{tabularx}
  \vspace{-0.3cm}
\end{table}


\noindent\textbf{Generalization.}
We test generalization on two axes: transfer to an unseen corpus, and robustness
to limited resources or reduced native training data.

\textit{Unseen data.} Table~\ref{tab:l2arctic_summary} reports transfer to L2-ARCTIC. Zero-shot transcript-guided features remain predictive, with DTW distance and mismatch rate showing consistent negative correlations.
A Ridge model trained on SpeechOcean762 transfers without retraining  (Acc./Flu./Pron. Score\ PCC$=0.506/0.492/0.526$), and a light calibration on L2-ARCTIC
further improves to $0.527/0.519/0.557$.


\textit{Limited native data.} Table~\ref{tab:main} compares the same system when the codebook, TLM, and Text2DUnit are trained on either $\sim$960 h or only $\sim$100 h of LibriSpeech. The combined system remains highly stable despite an order-of-magnitude reduction in native training data: accuracy is $0.668$ with $\sim$100 h and $0.661$ with $\sim$960 h, while the larger corpus yields only modest gains in fluency and prosody ($0.763/0.753$ vs.\ $0.757/0.748$). These results suggest that the framework does not require large native corpora to remain effective, making it well suited to low-resource settings where native ASR data is limited.

\section{Related Work}
\label{sec:related}

Automated pronunciation assessment has traditionally relied on GoP from forced-aligned ASR models~\cite{witt2000phone}, with later work extending this through end-to-end mispronunciation detection, alignment-aware training, phonetic/acoustic cue modeling, and Transformer-based regressors such as GOPT~\cite{leung2019cnn,9746727,10022472,gong2022transformer}. These methods typically require phoneme inventories, forced alignment, and labeled learner data. Recent SSL-based, multilingual, pseudo-labeling, and augmentation-based approaches reduce annotation cost~\cite{kim2022automatic,10022486,yang2022improving,lin2023multi,el2023multi,mixup,text-aug,zhang2022l2,Korzekwa2020DetectionOL,elkheir2023speechblender,el2024l1}, but still often depend on aligned L2 speech, learner supervision, or phone-level resources~\cite{kheir-etal-2023-automatic}. The closest zero-shot work, Liu et al.~\cite{liu2023zeroshot}, recovers masked HuBERT token spans and requires two forward passes per utterance. We instead use a single forward pass and an $n$-gram query, reducing computational cost while remaining competitive. 
Lee and Glass~\cite{lee2012comparison} introduced comparison-based methods that aligns learner and native speech with DTW  removing the need for per-phone labels; later works extended it over continuous embeddings~\cite{richter2023relative,lo2024zero}, and related efforts target children's and low-resource assessment~\cite{plantinga2019towards,smit2025towards}.

Our approach combines native-trained discrete-token surprisal with DTW alignment in the same discrete space, requiring no forced alignment, phoneme inventory, or learner labels.

\vspace{-0.2cm}

\section{Conclusion}
\label{sec:conclusion}

We presented a lightweight pronunciation assessment framework that learns from standard native speech resources and is designed for settings with little or no labeled learner data. The approach combines discrete-token surprisal from a native token language model with optional transcript-guided Text2DUnit--DTW alignment in the same discrete space. Experiments on SpeechOcean762 show that transcript guidance substantially improves over audio-only scoring and achieves competitive performance. The zero-retraining transfer to L2-ARCTIC further provides encouraging evidence of generalization. Although our current evaluation focuses on non-native English, the native-only training design makes the framework a promising direction for low-resource and non-English settings. Future work will test these settings directly and explore stronger text-to-unit models and calibration strategies under limited supervision.

\section{Use of Generative AI}
Generative AI tools were used during the preparation of this manuscript to
assist with language editing, grammar correction, and improving clarity of the
written text. These tools were not used to generate experimental results,
design the methodology, analyze data, or produce data or tables. All
technical content, experimental design, implementation, and interpretation of
results were carried out by the authors. The authors reviewed and edited all
AI-assisted text and take full responsibility for the final content of the
paper.
\section{Acknowledgement}

The work is supported by HBKU flagship research grant (HBKU-INT-VPR-FRG-03-09). The findings achieved herein are solely the responsibility of the authors.

\bibliography{mybib}
\bibliographystyle{IEEEtran}

\end{document}


\maketitle

\section{Tokenizer Selection Ablations}

This supplementary reports the full ablations used to select the discrete tokenizer in the main paper. All three studies are unsupervised, with no regression and no learner labels. For each configuration we compute the mean token surprisal on the SpeechOcean762 test set and measure its correlation with the human accuracy score, reported as Spearman $\rho$ and Pearson $r$. The correlations are negative because higher surprisal indicates greater phonotactic deviation, so a more negative value reflects a stronger label-free signal. We vary one factor at a time around a reference configuration of HuBERT, Layer~9, and $K{=}1024$, and the main paper takes the best value of each factor. 

\noindent\textbf{Encoder} (Table~\ref{tab:ablation_encoder}). Among four SSL encoders(HuBERT~\cite{hsu2021hubert}, WavLM~\cite{chen2022wavlm}, XLS-R~\cite{babu2021xls}, wav2vec~\cite{baevski2020wav2vec}), HuBERT base gives the strongest correlation ($\rho{=}-0.2584$). WavLM base is weaker, and XLS-R~300M and wav2vec~2.0 are near zero.

\noindent\textbf{Codebook size} (Table~\ref{tab:ablation_k}). The signal peaks at $K{=}512$ ($\rho{=}-0.2729$), and both smaller and larger codebooks are weaker, so we adopt $K{=}512$ in the main paper. Small codebooks ($K{=}100$) under-segment the acoustic space, and large ones ($K{=}2048$) fragment it and dilute the phonotactic statistics the TLM depends on.

\noindent\textbf{Layer} (Table~\ref{tab:ablation_layer}). For HuBERT, Layer~9 is clearly best ($\rho{=}-0.2584$), above Layers~6 and~12. This is consistent with prior layer-wise analyses, where middle HuBERT layers hold the most phonetic
information.
These ablations fix the final tokenizer as HuBERT base, Layer~9, and $K{=}512$, the configuration used throughout the main paper.

\begin{table}[H]
  \caption{Encoder ablation on SpeechOcean762 using Layer 9, $K=1024$, and a 3-gram TLM.}
  \label{tab:ablation_encoder}
  \centering
  \setlength{\tabcolsep}{8pt}
  \begin{tabular}{lcc}
    \toprule
    Encoder & Spearman $\rho$ & Pearson $r$ \\
    \midrule
    HuBERT base   & $-$0.2584 & $-$0.2141 \\
    WavLM base    & $-$0.1728 & $-$0.1419 \\
    XLS-R 300M    & $-$0.0702 & $-$0.0306 \\
    wav2vec~2.0   & $-$0.0087 & \phantom{$-$}0.0156 \\
    \bottomrule
  \end{tabular}
\end{table}

\begin{table}[H]
  \caption{Codebook size ablation on SpeechOcean762 using HuBERT Layer 9 and a 3-gram TLM.}
  \label{tab:ablation_k}
  \centering
  \setlength{\tabcolsep}{10pt}
  \begin{tabular}{rcc}
    \toprule
    $K$ & Spearman $\rho$ & Pearson $r$ \\
    \midrule
    100  & $-$0.1701 & $-$0.1448 \\
    256  & $-$0.2224 & $-$0.1811 \\
    512  & $-$0.2729 & $-$0.2267 \\
    1024 & $-$0.2584 & $-$0.2141 \\
    2048 & $-$0.2441 & $-$0.2111 \\
    \bottomrule
  \end{tabular}
\end{table}

\begin{table}[H]
  \caption{Layer ablation on SpeechOcean762 using HuBERT and $K=1024$ with a 3-gram TLM.}
  \label{tab:ablation_layer}
  \centering
  \setlength{\tabcolsep}{12pt}
  \begin{tabular}{rcc}
    \toprule
    Layer & Spearman $\rho$ & Pearson $r$ \\
    \midrule
    6  & $-$0.1650 & $-$0.1348 \\
    \textbf{9}  & $-$\textbf{0.2584} & $-$\textbf{0.2141} \\
    12 & $-$0.1559 & $-$0.1020 \\
    \bottomrule
  \end{tabular}
\end{table}
\bibliography{mybib}
\bibliographystyle{IEEEtran}